\documentclass[10pt,twocolumn,letterpaper]{article}

\usepackage{cvpr}              % To produce the CAMERA-READY version
\usepackage{booktabs}
\usepackage[dvipsnames]{xcolor}
\usepackage{colortbl}

\usepackage[dvipsnames]{xcolor}

\definecolor{darkred}{rgb}{0.6, 0.1, 0.05}
\definecolor{my_purple}{rgb}{0.74, 0.31, 1.0}
\definecolor{my_orange}{rgb}{1, 0.6, 0.0}
\definecolor{my_blue}{rgb}{0, 0.69, 0.94}
\definecolor{my_yellow}{rgb}{1, 0.75, 0.0}

\newcommand{\SupplementaryMaterial}{{\color{darkred} supplementary material}}
\newcommand{\SupplementaryVideo}{{\color{darkred} supplementary video}}

\usepackage{amsmath,amsfonts,bm}

\def\eqref#1{equation~\ref{#1}}
\def\1{\bm{1}}

\def\rvc{{\mathbf{c}}}
\def\rvd{{\mathbf{d}}}

\def\rvh{{\mathbf{h}}}

\def\rvo{{\mathbf{o}}}

\def\rvr{{\mathbf{r}}}

\def\rvx{{\mathbf{x}}}

\def\ervd{{\textnormal{d}}}

\def\rmA{{\mathbf{A}}}

\DeclareMathAlphabet{\mathsfit}{\encodingdefault}{\sfdefault}{m}{sl}
\SetMathAlphabet{\mathsfit}{bold}{\encodingdefault}{\sfdefault}{bx}{n}

\def\gU{{\mathcal{U}}}

\definecolor{cvprblue}{rgb}{0.21,0.49,0.74}
\usepackage[pagebackref,breaklinks,colorlinks,citecolor=cvprblue]{hyperref}

\def\paperID{1301} % *** Enter the Paper ID here
\def\confName{CVPR}
\def\confYear{2024}

\title{Rethinking Directional Integration in Neural Radiance Fields}

\author{Congyue Deng$^1$ \quad Jiawei Yang$^2$ \quad Leonidas Guibas$^1$ \quad Yue Wang$^{2}$ \\
$^1$Stanford University \quad $^2$University of Southern California \\
{\small $^1$\texttt{\{congyue, guibas\}@stanford.edu},~~ \texttt{$^2$\{yangjiaw, yue.w\}@usc.edu}}
}

\begin{document}
\maketitle
\begin{abstract}
Recent works use the Neural radiance field (NeRF) to perform multi-view 3D reconstruction, providing a significant leap in rendering photorealistic scenes. However, despite its efficacy, NeRF exhibits limited capability of learning view-dependent effects compared to light field rendering or image-based view synthesis.
To that end, we introduce a modification to the NeRF rendering equation which is as simple as a few lines of code change for any NeRF variations, while greatly improving the rendering quality of view-dependent effects.
By swapping the integration operator and the direction decoder network, we only integrate the positional features along the ray and move the directional terms out of the integration, resulting in a disentanglement of the view-dependent and independent components.
The modified equation is equivalent to the classical volumetric rendering in ideal cases on object surfaces with Dirac densities. Furthermore, we prove that with the errors caused by network approximation and numerical integration, our rendering equation exhibits better convergence properties with lower error accumulations compared to the classical NeRF.
We also show that the modified equation can be interpreted as light field rendering with learned ray embeddings.
Experiments on different NeRF variations show consistent improvements in the quality of view-dependent effects with our simple modification.
Project website: \url{https://cs.stanford.edu/~congyue/linerf/}.
\end{abstract}    
\section{Introduction}
\label{sec:intro}

\begin{figure}[t]
    \centering
    \includegraphics[width=\linewidth]{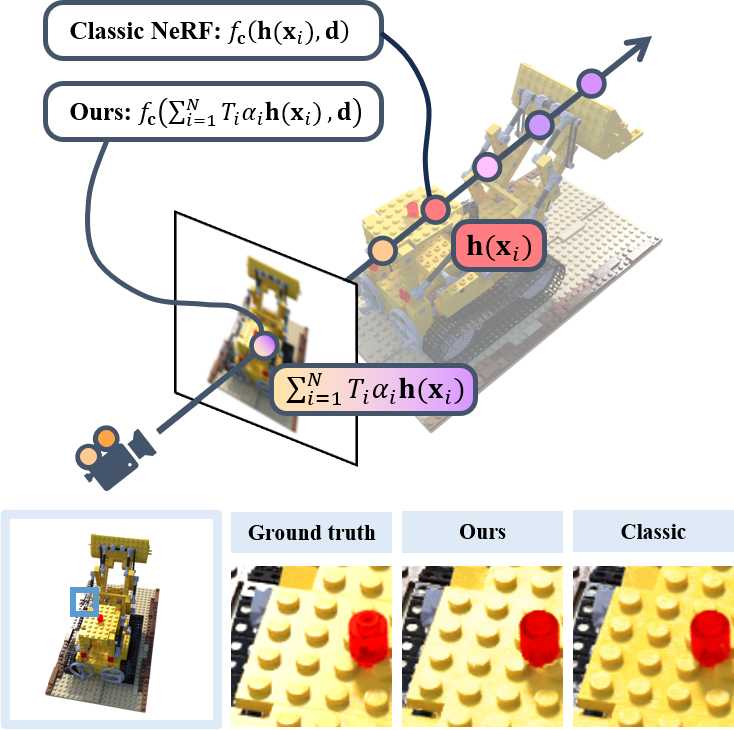}
    \caption{\textbf{Overview.}
    We introduce a modification to the NeRF rendering equation by swapping the integration operator and the direction decoder network, resulting in a disentanglement of the view-dependent and independent components.
    As simple as a few lines of code for any NeRF variations, our method can greatly improve the rendering quality of view-dependent effects.
    }
    \label{fig:teaser}
\end{figure}

% Background
Given a set of images sampled discretely around a scene, the task of reconstructing the original geometry and synthesizing novel views is a long-standing problem in computer vision and graphics \cite{seitz1996view, shade1998layered, shum2000review, chen2023view}.
At a cursory glance, the problem may be misconstrued as mere interpolation among densely sampled reference views; however,  synthesizing realistic and authentic unseen views remains a daunting task. This problem boils down to an intricate understanding of pixel-level correlations, structural image coherence, and the underlying 3D geometry. 

Recent advances in learning-based view synthesis have shown great success in representing scenes with neural networks.
Recent strides, such as NeRFs \cite{mildenhall2021nerf} and its successors \cite{lindell2021autoint, oechsle2021unisurf, barron2021mip, muller2022instant}, learn a 5D radiance field which maps 3D point locations and 2D view directions to corresponding opacities and colors. By learning the underlying 3D geometry, NeRF excels in predicting novel views of objects. In doing so, NeRF incorporates view-dependent components into its parameterization. However, they are only included in the last few layers of the neural network,  sacrificing its expressiveness in modeling non-Lambertian effects \cite{suhail2022light}.
Another line of work directly learns the radiance values in the pixel space \cite{wizadwongsa2021nex, wang2021ibrnet, attal2022learning, suhail2022light}, with light field rendering \cite{attal2022learning, suhail2022light} being their representatives.
These works show better performance in capturing view-dependent effects than NeRF. However, due to a lack of 3D awareness, they need additional learning of geometric inductive biases for regularization and usually need to maintain an explicit representation of pixel values such as the reference-view images even at inference time.

% Summarize our method and contributions
In this paper, we approach this problem from a different perspective by identifying a key issue with NeRF rendering: the existence of a large number of redundant view-direction queries. This redundancy arises because, while the spherical distribution of radiance is only physically meaningful on the object surfaces, the radiance function is learned throughout the entire 3D space (e.g., a view direction is even input to the network at empty space), over-consuming the network capacity as well as hindering the optimization process.

Inspired by this observation, we propose a simple yet effective framework that connects the volume rendering for NeRF with the ray-space embedding for light field rendering.
Viewed as a NeRF, it renders the intermediate positional features of the networks rather than the final radiance prediction.
Viewed as a light field, the rendered positional features provide a well-behaved embedding for image space pixels with epipolar correspondences, which endows the light field representation of 3D geometric inductive biases.
We name our modified framework as \textbf{LiNeRF}, for its dual interpretations as both a \emph{radiance field} and a \emph{light field}.

More concretely, our simple modification disentangles the positional and directional components, which efficiently enhances the rendering quality of view-dependent effects.
Specifically, we swap the order of two operators: the color prediction network and the integration operator along the ray.
That is, given a query point in 3D, we first compute its positional feature, the same as in NeRF. Then, instead of predicting the radiance at this point \wrt the ray direction, we directly compute the integral on the positional features along the ray, resulting in an aggregated view-independent feature at the ray starting point.
This integrated feature is further decoded together with the ray direction by the color prediction network, providing a radiance value for the whole ray (Fig. \ref{fig:teaser}).

In our framework, all networks and operations remain consistent with the classic NeRF rendering except for the order of applying these two operators. Albeit being less intuitive at first glimpse, the overall rendering equation is equivalent to classic NeRF in the ideal case with accurate density and color values.
More importantly, we show that with the errors inevitably introduced by the numerical integration and network predictions, our equation is a better estimator of the true radiance by having a tighter upper bound on the second-order term of its function approximation errors.

% Experiments
Our formulation can be easily incorporated into any existing NeRF framework with as simple as a few lines of code. Despite its simplicity, it can significantly enhance the rendering quality of view-dependent effects. We evaluate our method on several datasets with a variety of non-Lambertian surface materials, where we observe consistent improvements across all NeRF variations both visually and quantitatively.

To summarize, our key contributions are:
\begin{itemize}
    \item We propose a modification to the NeRF rendering equations that bridges the understanding of radiance fields and light fields.
    \item We provide theoretical proof that our formulation is a better numerical estimator of the radiance integration.
    \item Our method can be seamlessly integrated into any NeRF framework with only a few lines of code change.
    \item Experiments with different NeRF variations show consistent improvements in the rendering quality of view-dependent effects.
\end{itemize}
\section{Related Work}
\label{sec:related_work}

\paragraph{Neural radiance fields} 
Neural Radiance Field (NeRF) \cite{mildenhall2021nerf} has transformed the fields of 3D reconstruction and view synthesis, inspiring various extensions to enhance its original design \cite{verbin2022ref,barron2021mip,barron2023zip,barron2022mip,hedman2018deep,chen2023mobilenerf,reiser2023merf,muller2022instant}. Among these extensions, recent works \cite{hedman2021baking, chen2023mobilenerf, reiser2023merf} have adopted a strategy of rendering a feature map before decoding RGB output to improve the rendering speed. While conceptually similar to our method, these approaches typically embed view dependency directly into the feature maps during the rendering stage, which can limit the effective capture of view-dependent effects. Our approach, in contrast, explicitly disentangles view-dependent terms from the rendering process. This marks an important step beyond just efficiency improvements. In addition, our approach echoes the deferred shading process in 3D computer graphics \cite{hargreaves2004deferred}, wherein shading calculations are performed after the pixels are rendered, thereby efficiently computing complex lighting effects. However, our method takes this concept further by rendering a positional feature descriptor rather than a diffuse color, which greatly enhances the realism and accuracy of view-dependent effects.

\paragraph{Light field rendering} 
Introduced in \cite{levoy2023light}, the light field is a 4D function that specifies the radiance of any given ray traveling through the free space.
According to the Fourier slicing theorem, an RGB image is a 2D slice of the 4D light field in the frequency domain \cite{ng2005fourier}.
Due to this dimensionality gap, reconstructing a light field from multiview RGB images is non-trivial, instead, it either requires professional imaging devices such as the light field camera \cite{wilburn2001light, vaish2004using, wilburn2005high, ng2005light}, or relies on specific scene assumptions \cite{levin2010linear, shi2014light}.
Recent works \cite{buehler2023unstructured, kalantari2016learning, sitzmann2021light, srinivasan2017learning, attal2022learning, suhail2022light} combines learning techniques with light field rendering. However, they either require dense input sampling \cite{kalantari2016learning}, have a limited range of motion \cite{srinivasan2017learning}, are limited to simple scenes \cite{sitzmann2021light}, or cannot discard the reference views even at inference time \cite{suhail2022light}.

\paragraph{Image-based view synthesis}
Built on the notions that novel views can be rendered by interpolating the pixel values from a given set of input images, imaged-based view synthesis has been long studied in the computer graphics history \cite{seitz1996view, shade1998layered, shum2000review, chen2023view, laveau19943, seitz1995physically}.
Modern works \cite{chaurasia2013depth, penner2017soft, choi2019extreme, thies2018ignor, wizadwongsa2021nex, wang2021ibrnet} has incorporated deep learning techniques to predict depth information \cite{chaurasia2013depth, penner2017soft, choi2019extreme} or image features \cite{wizadwongsa2021nex, wang2021ibrnet} to facilitate pixel matching across different views.
Specifically, this line of method has shown its excellence in modeling view-dependent effects \cite{thies2018ignor, wizadwongsa2021nex, wang2021ibrnet}.
However, operating in the 2D pixel space, image-based representations usually lack a global understanding of the 3D scene and require additional inductive biases brought by geometric priors.

\section{Method}
\label{sec:method}

In this section, we present our simple yet effective modification to the NeRF volume rendering equation. We start by introducing the preliminaries of NeRF representation and rendering, and the notations and definitions will be used throughout the rest of the paper (Sec. \ref{sec:method:preliminaries}).
Then, we propose our modified rendering equation, which can also be interpreted as light field rendering (Sec. \ref{sec:method:ours}).
While being equivalent to classical NeRF in ideal cases, our modified rendering is proven to have better convergence properties under numerical integration (Sec. \ref{sec:method:taylor}).

\subsection{Preliminaries}
\label{sec:method:preliminaries}

\paragraph{Neural radiance field}
A neural radiance field (NeRF) represents a continuous scene as a 5D vector-valued function with a neural network $F_\Theta: (\rvx,\rvd)\to(\rvc,\sigma)$, whose input is a 3D location $\rvx = (x,y,z)$ and a 2D viewing direction $\rvd = (\theta,\varphi)$, and whose output is an emitted color $\rvc = (r,g,b)$ and a volumetric density $\sigma$.
In practice, it is usually implemented by two separate networks: the density network $f_\sigma(\rvh(\rvx))$ depending only on $\rvx$, and a color network $f_\rvc(\rvh(\rvx),\rvd)$ depending on both $\rvx$ and $\rvd$. Here, $\rvh(\rvx)$ is an input position feature encoding network\footnote{To avoid confusion, in the rest of the paper, we will call $\rvh(\rvx)$ the ``positional feature'' and the sinusoidal encoding on $\rvx$ the ``frequency encoding'' instead of the ``positional encoding (PE)'' as in most literature.}. During training, the parameters in $f_\rvc, f_\sigma, \rvh$ are jointly optimized.

\paragraph{Volume rendering with NeRF}
Given a camera ray $\rvr(t) = \rvo + t\rvd$ traveling through the radiance field with near and far bounds $t_n, t_f$, its expected color $C(\rvr)$ is
\begin{align}
    C(\rvr)
    &= \int_{t_n}^{t_f} T_t \sigma(\rvx_t) \rvc(\rvx_t, \rvd) \ervd t, \qquad  \\
    &= \int_{t_n}^{t_f} T_t f_\sigma(\rvh(\rvx_t)) f_\rvc(\rvh(\rvx_t), \rvd) \ervd t
\end{align}
where $\rvx_t$ denotes the points on the ray and $T_t = \exp\left( -\int_{t_n}^t \sigma(\rvr_s) \ervd s \right)$ is the accumulated transmittance along the ray.

To numerically estimate this integral, one can partition $[t_n, t_f]$ into $N$ bins and draw a point sample from each bin
\begin{equation}
    t_i \sim \gU\left[ t_n + \frac{i-1}{N}(t_f-t_n), t_n + \frac{i}{N}(t_f-t_n) \right]
\end{equation}
Here we adopt the evenly-spaced partitions and uniform point samples for simplicity, but our derivations do not rely on this assumption and can easily generalize to other sampling methods.

With the sampled points, the integral $C(\rvr)$ can be estimated differentiably with the quadrature rule \cite{max1995optical}
\begin{equation}
    \hat{C}(\rvr) = \sum_{i=1}^N T_i (1-\exp(-\sigma_i\delta_i)) \rvc_i
\end{equation}
where $\delta_i = t_{i+1} - t_i$ and $\text{where}~ T_i = \exp \left( -\sum_{j=1}^{i-1}\sigma_j\delta_j \right)$.
This is similar to the traditional alpha compositing with alpha values $\alpha_i = 1 - \exp(-\sigma_i\delta)$.

\paragraph{Light field rendering}
Other than the radiance fields that model the volumetric radiances across the 3D space, a light field directly represents the color observed along each ray direction, which is a function $L: \rvr\to\rvc$ on the space of oriented lines that associate a radiance value to each ray.

According to the Fourier Slicing Theorem, a 2D image captured by an RGB camera is a 2D slice of the 4D light field \cite{ng2005fourier}, and this dimensionality gap makes directly reconstructing the light field extremely challenging without further constraints.
Recent learning-based works address this problem by learning efficient ray embeddings \cite{attal2022learning, suhail2022light}. Specifically, one could incorporate geometric inductive biases using the epipolar geometry, where two rays $\rvr^1, \rvr^2$ have higher feature similarities if they intersect with the object surface at the same point.

\subsection{Our Solution}
\label{sec:method:ours}

\begin{figure}[t]
    \centering
    \includegraphics[width=.85\linewidth]{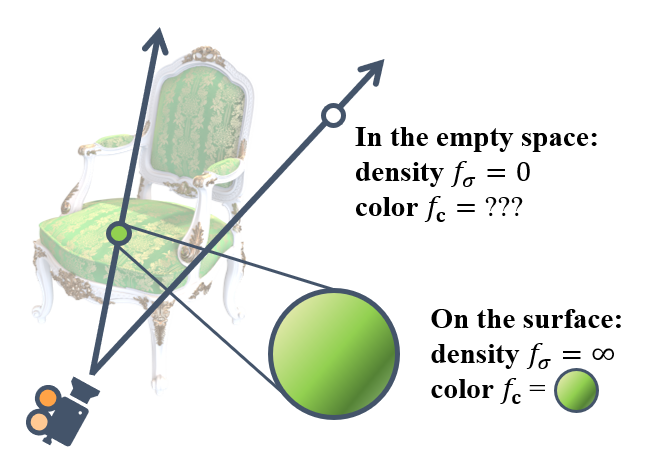}
    \caption{\textbf{Ill-defined queries of view-dependent radiances in the free space.}
    In NeRF representations, all networks $f_\rvc, f_\sigma, \rvh$ are defined over the entire space. 
    But for points off the object surface, only the positional feature $\rvh(\rvx)$ and the density network $f_\sigma(\rvh(\rvx))$ have physical meanings, while the view-dependent radiance network $f_\rvc(\rvh(\rvx),\rvd)$ is dummy.
    }
    \label{fig:intuition}
\end{figure}

To enable efficient training and complete differentiability, all NeRF networks $f_\rvc, f_\sigma, \rvh$ are defined over the entire space. While this is natural for the positional feature $\rvh(\rvx)$ and the density network $f_\sigma(\rvx)$, the view-dependent radiance network $f_\rvc(\rvx,\rvd)$ only have physical meanings for surface points but is ill-defined for the empty-space points with zero densities (Fig. \ref{fig:intuition}).
This means that learning $f_\rvc$ over the entire 3D space is over-consuming the network representation power.

To avoid such redundancy and encourage $f_\rvc$ to focus more on the view-dependent radiance distributions \emph{on the object surface}, we propose a simple modification of the rendering equation as
\begin{equation}
    C'(\rvr)
    = f_\rvc\left( \int_{t_n}^{t_f} T_t f_\sigma\left(\rvh(\rvx_t)\right) \rvh(\rvx_t)  \ervd t, \rvd \right).
\end{equation}
Instead of obtaining the view-dependent radiances $f_\rvc(\rvh(\rvx_t), \rvd)$ for $\rvx_t$ and integrating them along the ray, we directly integrate the positional features $\rvh(\rvx_t)$, resulting in an aggregated feature at the near-camera point $\rvx_{t_n}$. We then directly obtain the overall radiance of the ray by decoding this integrated feature through $f_\rvc$ together with the view direction $\rvd$ (Fig. \ref{fig:teaser}).
In other words, we swap the order of the integration operator and $f_\rvc$ in the rendering equation.

While being extremely simple, this modification disentangles the positional components and view-dependent components in the rendering equation. The positional networks $f_\sigma(h(\rvx))$ and $h(\rvx)$ are queried for all points along the ray, but the radiance network $f_\rvc(h(\rvx))$ and $h(\rvx)$ with $\rvd$ input is only called once for each ray, avoiding the physically meaningless queries for the off-surface points which wastes the network capacity.
We provide more theoretical explanations and experimental evidence in the following sections.

\paragraph{Equivalence to NeRF}
For object surfaces with alpha value 1, there is one unique point $\rvx_* = \rvo + t_*\rvd$ where the ray $\rvr$ first intersects with the surface. In ideal cases with ground-truth radiance distributions, the density function $f_\sigma$ is a Dirac delta function at $\rvx_*$ and $f_\rvc(\rvh(\rvx_*), \rvd)$ is the surface color. Therefore, both integrals degenerate to
\begin{equation}
    C(\rvr) = C'(\rvr) = f_\rvc((\rvh(\rvx_*), \rvd),
\end{equation}
implying an equivalence between the classic and our modified rendering equation.

\paragraph{Interpreted as light field rendering}
From another perspective, our rendering equation can also be interpreted as a light field $L(\rvr) = f_\rvc(H(\rvr), \rvd)$, with ray embedding
\begin{equation}
    (H(\rvr), \rvd), ~\text{where}~
    H(\rvr) = \int_{t_n}^{t_f} T_t f_\sigma(\rvh(\rvx_t)) \rvh(\rvx_t)  \ervd t
\end{equation}
For ground-truth radiance fields, $H(\rvr) = h(\rvx_*)$ for any ray $\rvr$ intersecting with the object surface at $\rvx_*$ for the first time, and thus the ray embedding becomes $(h(\rvx_*), \rvd)$. This endows the light field representation with view consistency by having similar embeddings for the corresponding epipolar points from different image views, which is exactly the geometric inductive bias we desire in light field rendering.

\subsection{Numerical Integration}
\label{sec:method:taylor}

We employ the same point sampling and numerical integration as in the NeRF volume rendering to compute the integral on $\rvh(\rvx_t)$ with alpha compositing based on the predicted densities, which then gives the color estimation
\begin{equation}
    \hat{C}'(\rvr) = f_\rvc\left( \sum_{i=1}^N T_i (1-\exp(-\sigma_i\delta_i)) \rvh(\rvx_i), \rvd \right).
\end{equation}

\paragraph{Alpha blending with background}
Let $\lambda_i = T_i(1-\exp(-\sigma_i\delta_i))$ be the scalar weights on the sampled points $\rvx_i$, the accumulated alpha value for the foreground is $\Lambda_{\text{fg}} = \sum_{i=1}^N \lambda_i$.
When $\Lambda_{\text{fg}} < 1$, we apply a normalization to the weights as $\hat{\lambda_i} = \lambda_i / \Lambda_{\text{fg}}$ and compute the feature integration with the normalized weights $\sum_{i=1}^N \hat{\lambda_i} \rvh(\rvx_i)$. This is for eliminating the scaling effects in the feature space and maintaining its stability during optimization.
In the end, we blend the foreground and background with
\begin{equation}
    \Lambda_{\text{fg}} f_\rvc\left( \sum_{i=1}^N \frac{\lambda_i}{\Lambda_{\text{fg}}} \rvh(\rvx_i), \ervd \right)
    + (1 - \Lambda_{\text{fg}}) C_{\text{bg}}
\end{equation}
where $C_{\text{bg}}$ is the background color which can be either a constant color or a function on $\rvd$.
Note that the foreground radiance is weighted by $\Lambda_{\text{fg}}$ in the linear sum, as compensation for the scaling applied to the per-point weights.
This doesn't exist in the classic NeRF foreground-background composition because the per-point weights for integration $\lambda_i/\Lambda_{\text{fg}}$ are out of $f_\rvc$ and thus cancel with the overall foreground weighting $\Lambda_{\text{fg}}$.

\paragraph{A better radiance approximator}
While we show that in ideal cases $C(\rvr) = C'(\rvr)$, errors are inevitable in both network predictions and the numerical integration.
Now we show that under more realistic conditions, our rendering equation is a better approximation of the ground-truth radiance.

For simplicity, we write $\rvh(\rvx_i) = \rvh_i$.
The rendering equations for the classical NeRF and our modified version can be written as
\begin{align}
    \text{NeRF:}\quad \hat{C}(\rvr)
    &= \sum_{i=1}^N \lambda_i f_\rvc(\rvh_i, \rvd) \label{C}\\
    \text{Ours:}\quad \hat{C}'(\rvr)
    &= f_\rvc\left(\sum_{i=1}^N \lambda_i\rvh_i, \rvd\right) \label{C_square}
\end{align}
Here we assume that for both rendering equations, the 3D fields are represented by the same networks $f_\sigma, f_\rvc, \rvh$, and the point samplings are also the same.

As before, we assume that ray $\rvr$ intersects with the object surface at point $\rvx_*$, giving a ground-truth feature $\rvh_* = \rvh(\rvx_*)$ and we can write $\rvh_i = \rvh_* + \Delta\rvh_i$.
When the networks are trained to convergence, we have non-zero densities centered around $\rvx_*$, which means that for $\sigma_i \gg 0$, $\Delta\rvh_i \ll 1$.
Computing the Taylor expansions on $\hat{C}(\rvr)$ and $\hat{C}'(\rvr)$, their zero and first-order terms are both
\begin{equation}
    f_\rvc(\rvh_*,\rvd) + \nabla_\rvh f_\rvc(\rvh_*,\rvd) \sum_{i=1}^N \lambda_i \Delta\rvh_i
\end{equation}
However, the second-order terms are
\begin{align}
     T^{(2)}\hat{C}(\rvr) &=
     \frac{1}{2} \sum_{i=1}^N \lambda_i \Delta\rvh_i^t \nabla_\rvh^2 f_\rvc(\rvh_*,\rvd) \Delta\rvh_i \\
     T^{(2)}\hat{C}'(\rvr) &=
     \frac{1}{2} \sum_{i=1}^N \sum_{j=1}^N \lambda_i\lambda_j \Delta\rvh_i^t \nabla_\rvh^2 f_\rvc(\rvh_*,\rvd) \Delta\rvh_j
\end{align}
Let $u_i = \|\nabla_\rvh^2 f_\rvc(\rvh_*,\rvd)\|^{1/2} \|\Delta\rvh_i\|$ where $\|\nabla_\rvh^2 f_\rvc(\rvh_*,\rvd)\|$ is the matrix Euclidean norm, we can obtain (tight) upper bounds
\begin{align}
    \left| T^{(2)}\hat{C} \right| &\leqslant
    U\left(\left| T^{(2)}\hat{C} \right|\right) =
    \frac{1}{2} \sum_{i=1}^N \lambda_i u_i^2 \\
    \left| T^{(2)}\hat{C}' \right| &\leqslant
    U\left( \left| T^{(2)}\hat{C}' \right|\right) =
    \frac{1}{2} \left( \sum_{i=1}^N \lambda_i u_i \right)^2
\end{align}
Without loss of generality, we can assume that $\Lambda_{\text{fg}} = \sum_i \lambda_i = 1$ for the foreground surface with accumulated alpha value 1, and proofs for $\Lambda_{\text{fg}} < 1$ blended with constant background colors are similar.
By Jensen's inequality, we have
\begin{equation}
     U\left(\left| T^{(2)}\hat{C} \right|\right) \geqslant
     U\left(\left| T^{(2)}\hat{C}' \right| \right)
\end{equation}
indicating that $\hat{C}'(\rvr)$ is a better radiance approximator with a tighter error bound than $\hat{C}(\rvr)$.
Step-by-step derivations can be found in the \SupplementaryMaterial.

\section{Experiments}
\label{sec:exp}

In this section, we show that our modification brings consistent improvements in learning view-dependent rendering effects on different NeRF variations regardless of their network architectures (Sec. \ref{sec:exp:all_nerfs}). We also provide more qualitative and quantitative results on synthetic and real-captured datasets (Sec. \ref{sec:exp:more_comparisons}) with existing works curated for non-Lambertian view synthesis as references.

\paragraph{Datasets}
We mainly evaluate our method on the Shiny Blender dataset \cite{verbin2022ref} which consists of 6 different glossy objects rendered in Blender with non-Lambertian material properties. The camera setups and training/testing splits are similar to the Blender dataset \cite{verbin2022ref} with 100 training and 200 testing images of resolution $800\times800$.
In addition, we also evaluate our method on the 8 objects from the Blender dataset \cite{verbin2022ref}, where we address our superiority on the non-Lambertian surface regions.

Finally, we also test on the real-captured Shiny dataset \cite{wizadwongsa2021nex} comprising 8 forward-facing scenes with view-dependent effects. We use the same training/test split and image resolution as in prior works \cite{wizadwongsa2021nex, suhail2022light}.

\paragraph{Evaluation metrics}
We report the widely adopted view-synthesis metrics: peak signal-to-noise ratio (PSNR), structural similarity index measure (SSIM), and learned perceptual image patch similarity (LPIPS) with the VGG network backbone. We mainly compare the PSNR for per-pixel rendering errors which best measures the model capabilities of representing color variations across views.

\subsection{Results on Different NeRF Variations}
\label{sec:exp:all_nerfs}

\begin{table}[t]
    \centering
    \begin{tabular}{l|l|c|cc}
    \toprule
        Posi. enc. & View enc. & $\dim(\rvh)$ & Classic & Ours \\
    \midrule
        Sinusoidal & Sinusoidal & 256 & 30.53 & \textbf{30.68} \\
        Sinusoidal & SH & 256         & 30.16 & \textbf{30.64} \\
        TiledGrid & Sinusoidal & 15   & 29.34 & \textbf{29.85} \\
        TiledGrid & SH & 15           & 29.48 & \textbf{29.92} \\
        HashGrid & Sinusoidal & 15    & 29.36 & \textbf{29.75} \\
        HashGrid & SH & 15            & 29.62 & \textbf{30.36} \\
    \bottomrule
    \end{tabular}
    \caption{Results of our method with different NeRF variations on the Shiny Blender dataset \cite{verbin2022ref}.
    Our simple modification results in an average boost of PSNR by $+0.45$.
    Per-scene results can be found in the \SupplementaryMaterial.
    }
    \label{tab:exp_all_nerfs}
\end{table}

\begin{figure*}[t]
    \centering
    \includegraphics[width=\linewidth]{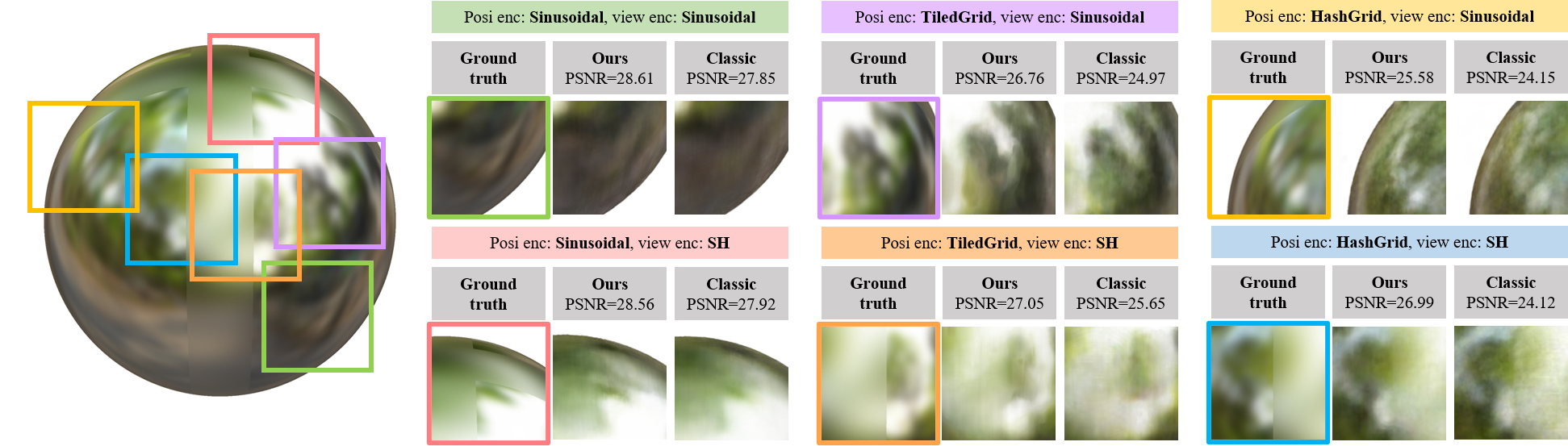}
    \caption{\textbf{Results of our method on different NeRF variations.}
    Here we highlight this challenging scene with a reflective metallic ball having two different roughness values distributed on its surface material.
    Our simple modification brings significant improvements to the rendered surface reflections with an average boost of PSNR by $+1.49$.
    More comparisons in pixel-aligned forms can be found in the \SupplementaryVideo.}
    \label{fig:exp_all_nerfs}
\end{figure*}

Table \ref{tab:exp_all_nerfs} shows the quantitative results of our modification on a set of NeRF variations with different network architectures and input encodings.
For the input positional feature network, we experiment with the vanilla MLP NeRF with sinusoidal frequency encodings \cite{mildenhall2021nerf}, as well as the latest grid-based NeRF architecture \cite{muller2022instant} with tiled grids or hash grids.
For view encoding, we test both the sinusoidal encoding and the spherical harmonics (SH) encoding.
$\dim(\rvh) = \dim(\rvh(\rvx_t))$ is the dimension of the positional feature to be integrated along the ray in our formulation, which is 256 for MLP-based NeRFs with Sinusoidal frequency encodings and 15 for grid-based NeRFs with multi-resolution grid encodings.
All network implementations stay exactly the same between the classic NeRF and our modified version, except for swapping the order of the two operators in the rendering equation.
As shown in the table, our simple modification gives rise to a consistent improvement in all these NeRF variations, with an average boost of PSNR by $+0.45$.

Figure \ref{fig:exp_all_nerfs} shows the qualitative results on a test view with all the different NeRF architectures.
Here we highlight one of the most challenging scenes with a reflective metallic ball having two different roughness values distributed on its surface material.
Our simple modification brings significant improvements to the faithfulness and detail clarity of the rendered surface reflections, with an average boost of PSNR by $+1.49$.
More qualitative comparisons can be found in the \SupplementaryVideo~ and are shown in pixel-aligned forms.

\subsection{More Results and Comparisons}
\label{sec:exp:more_comparisons}

\begin{figure*}[t]
    \centering
    \includegraphics[width=\linewidth]{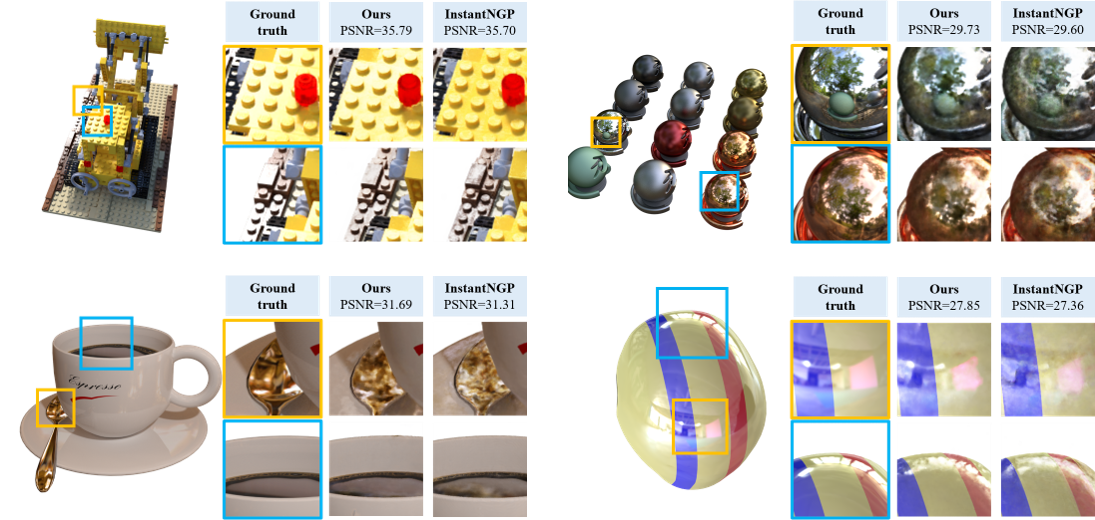}
    \caption{\textbf{Results on synthetic datasets.}
    \textbf{Top row:} Scenes from the Blender dataset \cite{mildenhall2021nerf}. 
    \textbf{Bottom row:} Scenes from the Shiny Blender dataset \cite{verbin2022ref}.
    More comparisons in pixel-aligned forms can be found in the \SupplementaryVideo.}
    \label{fig:exp_synthetic}
\end{figure*}

\begin{table}[t]
    \centering
    \begin{tabular}{l|ccc}
    \toprule
        Method & PSNR$\uparrow$ & SSIM$\uparrow$ & LPIPS $\downarrow$ \\
    \midrule
        InstantNGP \cite{muller2022instant} & 29.62 & 0.904 & 0.148 \\
        InstantNGP+Ours & \textbf{30.36} & \textbf{0.907} & \textbf{0.141} \\
    \midrule
        \rowcolor[gray]{0.9} PhySG \cite{zhang2021physg} & 26.21 & 0.921 & 0.121 \\
        \rowcolor[gray]{0.9} Mip-NeRF \cite{barron2021mip} & 29.76 & 0.942 & 0.092 \\
        \rowcolor[gray]{0.9} Ref-NeRF \cite{verbin2022ref} & 35.96 & 0.967 & 0.058 \\
    \bottomrule
    \end{tabular}
    \caption{Results on the Shiny Blender dataset \cite{verbin2022ref}.
    Per-scene results can be found in the \SupplementaryMaterial.
    }
    \label{tab:exp_refnerf}
\end{table}

\paragraph{Shiny Blender Dataset}
Table \ref{tab:exp_refnerf} shows our additional results on the Shiny Blender dataset \cite{verbin2022ref}.
We utilized the InstantNGP architecture with hash grid positional encodings and spherical harmonic view direction encodings for both the classic NeRF and our modified rendering equation.

\begin{figure*}[t]
    \centering
    \includegraphics[width=\linewidth]{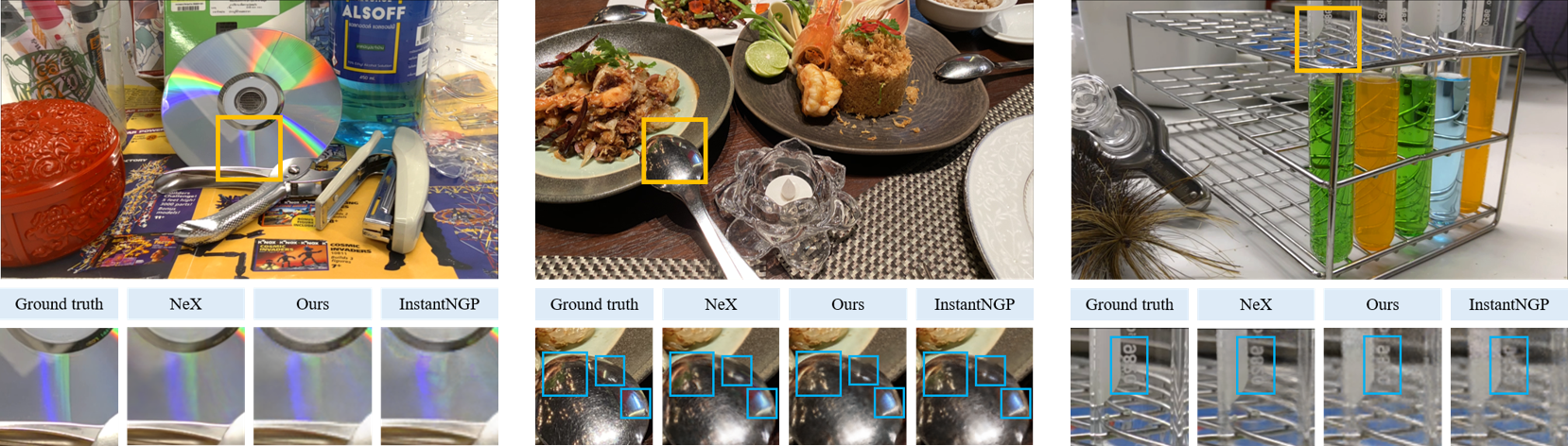}
    \caption{\textbf{Results on the Shiny dataset.}
    Our method can efficiently model a variety of view-dependent effects, such as the light interference on the CD (left), the reflections on the metal spoon (middle), or the refraction on the glass test tube (right).
    More comparisons in pixel-aligned forms can be found in the \SupplementaryVideo.}
    \label{fig:exp_shiny}
\end{figure*}

We also list the performances of other methods curated for surface BRDFs and view-dependent effects as references.
Although Ref-NeRF \cite{verbin2022ref} shows compelling results here, it is curated for reflective surfaces with reflected radiance predictions and diffuse/specular separations, instead of view-dependent effects in general (for example, the color interference in Fig. \ref{fig:exp_shiny} left, or the refractions in Fig. \ref{fig:exp_shiny} right). Moreover, it is at the cost of increased computation with slower inference and significantly slower training.

\begin{table}[t]
    \centering
    \begin{tabular}{l|ccc}
    \toprule
        Method & PSNR$\uparrow$ & SSIM$\uparrow$ & LPIPS $\downarrow$ \\
    \midrule
        InstantNGP \cite{muller2022instant} & 33.09 & \textbf{0.961} & 0.054 \\
        InstantNGP+Ours & \textbf{33.17} & \textbf{0.961} & \textbf{0.051} \\
    \midrule
        \rowcolor[gray]{0.9} NeRF \cite{mildenhall2021nerf} & 31.01 & 0.953 & 0.050 \\
        \rowcolor[gray]{0.9} IBRNet \cite{wang2021ibrnet} & 28.14 & 0.942 & 0.072 \\
        \rowcolor[gray]{0.9} Mip-NeRF \cite{barron2021mip} & 33.09 & 0.961 & 0.043 \\
        \rowcolor[gray]{0.9} LFNR \cite{suhail2022light} & 33.85 & 0.981 & 0.024 \\
        \rowcolor[gray]{0.9} Ref-NeRF \cite{verbin2022ref} & 33.99 & 0.966 & 0.038 \\
    \bottomrule
    \end{tabular}
    \caption{Results on the Blender dataset \cite{mildenhall2021nerf}.
    Per-scene results can be found in the \SupplementaryMaterial.
    }
    \label{tab:exp_blender}
\end{table}

\paragraph{Blender Dataset}
Table \ref{tab:exp_blender} shows our results on the Blender dataset \cite{mildenhall2021nerf}. We observe that our method shows comparable performances on the scenes mostly consisting of Lambertian surfaces but has more noticeable improvements on the non-Lambertian ones. Per-scene quantitative results can be found in the \SupplementaryMaterial.

Figure \ref{fig:exp_synthetic} shows the qualitative results on the two NeRF synthetic datasets. We can see that our formulation renders better non-Lambertian surface light conditions (such as the Lego on the top left and the coffee on the bottom left) and clearer details in the reflected environment maps (such as the materials ball on the top right, the metal spoon on the bottom left, and the helmet on the bottom right).

\begin{table}[t]
    \centering
    \begin{tabular}{l|ccc}
    \toprule
        Method & PSNR$\uparrow$ & SSIM$\uparrow$ & LPIPS $\downarrow$ \\
    \midrule
        InstantNGP \cite{muller2022instant} & 26.34 & \textbf{0.833} & \textbf{0.179} \\
        InstantNGP+Ours & \textbf{26.36} & \textbf{0.833} & \textbf{0.179} \\
    \midrule
        \rowcolor[gray]{0.9} NeRF \cite{mildenhall2021nerf} & 25.60 & 0.851 & 0.259 \\
        \rowcolor[gray]{0.9} NeX \cite{wizadwongsa2021nex} & 26.45 & 0.890 & 0.165 \\
        \rowcolor[gray]{0.9} IBRNet \cite{wang2021ibrnet} & 26.50 & 0.863 & 0.122 \\
        \rowcolor[gray]{0.9} LFNR \cite{suhail2022light} & 27.34 & 0.907 & 0.045 \\
    \bottomrule
    \end{tabular}
    \caption{Results on the Shiny dataset \cite{wizadwongsa2021nex}.
    Per-scene results can be found in the \SupplementaryMaterial.
    }
    \label{tab:exp_shiny}
\end{table}

\paragraph{Shiny Dataset}
Table \ref{tab:exp_shiny} shows our results on the Shiny dataset \ref{tab:exp_shiny}. Our PSNR is slightly lower than NeX \cite{wizadwongsa2021nex} and IBRNet \cite{wang2021ibrnet}, both of which are image-based rendering methods and store explicit pixel value representations.
Our improvements here are less significant compared to the previous two object-centric datasets, which, based on our observations, is partly because of challenges in complex scene representations dominating the evaluation metrics (examples are shown in the \SupplementaryMaterial).

Figure \ref{fig:exp_shiny} shows our qualitative results. We showcase different types of view-dependent effects, including light interferences (the CD on the left), reflection (the metal spoon in the middle), and refraction (the glass test tube on the right). Our method demonstrates visually noticeable improvements compared to the classic Instant NGP rendering.

\subsection{Feature Choices for Integration}
\label{sec:exp:feat_choice}

\begin{figure}[t]
    \centering
    \includegraphics[width=\linewidth]{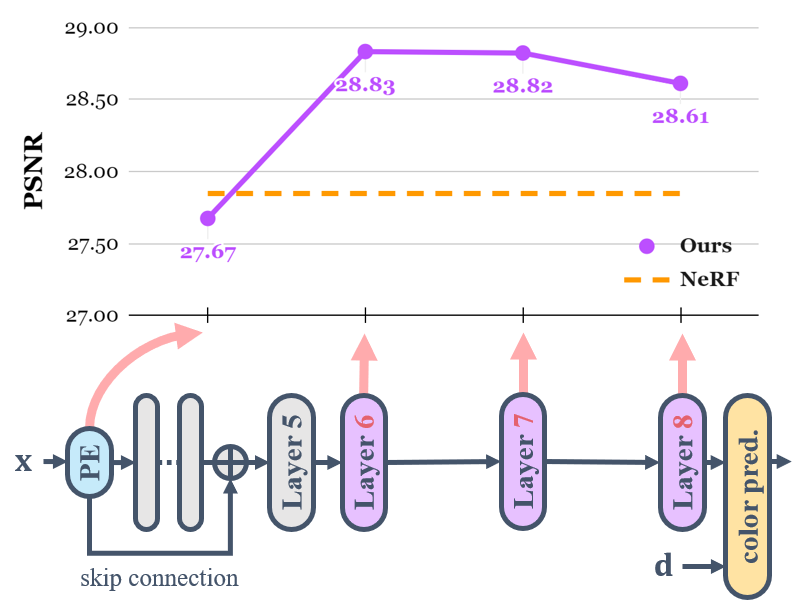}
    \caption{\textbf{Results of different features for integration.}
    We evaluate the PSNR on the ``ball'' scene from the Shiny Blender dataset \cite{verbin2022ref}.
    We experiment with an 8-layer MLP positional feature network as depicted at the bottom.
    The {\color{my_purple}purple dots} are for our rendering equation integrated with features from the sinusoidal encoding, MLP layers 6, 7, and 8 respectively.
    The {\color{my_orange}orange dashed line} is the result of classic NeRF which integrates the radiance values predicted together with $\rvd$.
    }
    \label{fig:exp_feat_choice}
\end{figure}

To further study the properties of our rendering equation, we test its variations by sending the features from different network layers into the integration along the ray.
Specifically, let $\rvh$ be the positional feature network and $f_\rvc$ be the radiance prediction network.
We can decompose $\rvh$ into $\rvh = \rvh_1 \circ \rvh_2$ with $\rvh_1, \rvh_2$ being different sequences of network layers.
We can then view $\rvh_2(\rvx)$ as the positional feature to be integrated, and $\rvh_1, f_\rvc$ together as the directional radiance prediction network $f_\rvc(\rvh_1(\cdot), \cdot)$.
The rendering equation in this sense becomes
\begin{equation}
    f_\rvc\left( \rvh_1\left( \int_{t_n}^{t_f} T_t f_\sigma(\rvh_2(\rvx_t)) \rvh_2(\rvx_t)  \ervd t \right), \rvd \right)
\end{equation}

We experiment with a vanilla NeRF architecture whose positional feature network is $\rvh(\rvx)$ is an 8-layer MLP with a sinusoidal frequency encoding at the beginning.
We take the features from the sinusoidal layer and the layers after the skip connection for integration along the ray and evaluate
The feature from the sinusoidal frequency encoding is of dimension 63, and the rest from MLP layers 6, 7, 8 are of dimension 256.

Figure \ref{fig:exp_feat_choice} shows the results of different network-layer choices for integration.
While directly integrating the sinusoidal encoding results in a performance drop, our integration with the MLP layer outputs is consistently better than classic NeRF which doesn't disentangle the $\rvx$ and $\rvd$ components. This further indicates our robustness over network architecture designs.
\section{Conclusions}
\label{sec:conclusions}

In this paper, we address the challenge of learning view-dependent effects in existing NeRF frameworks.
Based on the observation of the redundancy in directional queries, we propose a simple yet effective modification to the NeRF rendering equation.
Our formulation also builds a connection between the intuitions of NeRF volume rendering and light field rendering.
Theoretically, we show that our formulation is equivalent to classic NeRF in ideal cases with Dirac delta densities on the surface. In addition, with  inaccurate network prediction and numerical integration, ours is a better estimator for the overall radiance with a tighter upper bound on its function approximation error.
Experimentally, our modification can be easily incorporated into any existing NeRF frameworks and significantly improve the rendering quality of view-dependent effects with different NeRF variations.

\paragraph{Limitations and future work}
As a ``free lunch'' for any NeRF framework, our formulation brings a noticeable enhancement compared to the classic NeRF rendering equation.
However, compared to the image-based view synthesis methods curated for non-Lambertian effects which store explicit representations for pixel values, our improvements still remain limited.
How to combine the implicit radiance field rendering and explicit pixel-based rendering (such as light field rendering) more closely remains an interesting future direction.

Another future direction is to further study the choices of features for integration. As revealed in Section \ref{sec:exp:feat_choice}, integrating the features from different network layers can result in different performances within the same NeRF framework. We can explore strategies to choose the most appropriate features given network architectures. 
% How to choose the most appropriate features with given different network architectures could be a future work.

{
    \small
    \bibliographystyle{ieeenat_fullname}
    \bibliography{main}

\begin{thebibliography}{41}
\providecommand{\natexlab}[1]{#1}
\providecommand{\url}[1]{\texttt{#1}}
\expandafter\ifx\csname urlstyle\endcsname\relax
  \providecommand{\doi}[1]{doi: #1}\else
  \providecommand{\doi}{doi: \begingroup \urlstyle{rm}\Url}\fi

\bibitem[Attal et~al.(2022)Attal, Huang, Zollh{\"o}fer, Kopf, and Kim]{attal2022learning}
Benjamin Attal, Jia-Bin Huang, Michael Zollh{\"o}fer, Johannes Kopf, and Changil Kim.
\newblock Learning neural light fields with ray-space embedding.
\newblock In \emph{Proceedings of the IEEE/CVF Conference on Computer Vision and Pattern Recognition}, pages 19819--19829, 2022.

\bibitem[Barron et~al.(2021)Barron, Mildenhall, Tancik, Hedman, Martin-Brualla, and Srinivasan]{barron2021mip}
Jonathan~T Barron, Ben Mildenhall, Matthew Tancik, Peter Hedman, Ricardo Martin-Brualla, and Pratul~P Srinivasan.
\newblock Mip-nerf: A multiscale representation for anti-aliasing neural radiance fields.
\newblock In \emph{Proceedings of the IEEE/CVF International Conference on Computer Vision}, pages 5855--5864, 2021.

\bibitem[Barron et~al.(2022)Barron, Mildenhall, Verbin, Srinivasan, and Hedman]{barron2022mip}
Jonathan~T Barron, Ben Mildenhall, Dor Verbin, Pratul~P Srinivasan, and Peter Hedman.
\newblock Mip-nerf 360: Unbounded anti-aliased neural radiance fields.
\newblock In \emph{Proceedings of the IEEE/CVF Conference on Computer Vision and Pattern Recognition}, pages 5470--5479, 2022.

\bibitem[Barron et~al.(2023)Barron, Mildenhall, Verbin, Srinivasan, and Hedman]{barron2023zip}
Jonathan~T Barron, Ben Mildenhall, Dor Verbin, Pratul~P Srinivasan, and Peter Hedman.
\newblock Zip-nerf: Anti-aliased grid-based neural radiance fields.
\newblock \emph{arXiv preprint arXiv:2304.06706}, 2023.

\bibitem[Buehler et~al.(2023)Buehler, Bosse, McMillan, Gortler, and Cohen]{buehler2023unstructured}
Chris Buehler, Michael Bosse, Leonard McMillan, Steven Gortler, and Michael Cohen.
\newblock Unstructured lumigraph rendering.
\newblock In \emph{Seminal Graphics Papers: Pushing the Boundaries, Volume 2}, pages 497--504. 2023.

\bibitem[Chaurasia et~al.(2013)Chaurasia, Duchene, Sorkine-Hornung, and Drettakis]{chaurasia2013depth}
Gaurav Chaurasia, Sylvain Duchene, Olga Sorkine-Hornung, and George Drettakis.
\newblock Depth synthesis and local warps for plausible image-based navigation.
\newblock \emph{ACM Transactions on Graphics (TOG)}, 32\penalty0 (3):\penalty0 1--12, 2013.

\bibitem[Chen and Williams(2023)]{chen2023view}
Shenchang~Eric Chen and Lance Williams.
\newblock View interpolation for image synthesis.
\newblock In \emph{Seminal Graphics Papers: Pushing the Boundaries, Volume 2}, pages 423--432. 2023.

\bibitem[Chen et~al.(2023)Chen, Funkhouser, Hedman, and Tagliasacchi]{chen2023mobilenerf}
Zhiqin Chen, Thomas Funkhouser, Peter Hedman, and Andrea Tagliasacchi.
\newblock Mobilenerf: Exploiting the polygon rasterization pipeline for efficient neural field rendering on mobile architectures.
\newblock In \emph{Proceedings of the IEEE/CVF Conference on Computer Vision and Pattern Recognition}, pages 16569--16578, 2023.

\bibitem[Choi et~al.(2019)Choi, Gallo, Troccoli, Kim, and Kautz]{choi2019extreme}
Inchang Choi, Orazio Gallo, Alejandro Troccoli, Min~H Kim, and Jan Kautz.
\newblock Extreme view synthesis.
\newblock In \emph{Proceedings of the IEEE/CVF International Conference on Computer Vision}, pages 7781--7790, 2019.

\bibitem[Hargreaves and Harris(2004)]{hargreaves2004deferred}
Shawn Hargreaves and Mark Harris.
\newblock Deferred shading.
\newblock In \emph{Game Developers Conference}, page~31, 2004.

\bibitem[Hedman et~al.(2018)Hedman, Philip, Price, Frahm, Drettakis, and Brostow]{hedman2018deep}
Peter Hedman, Julien Philip, True Price, Jan-Michael Frahm, George Drettakis, and Gabriel Brostow.
\newblock Deep blending for free-viewpoint image-based rendering.
\newblock \emph{ACM Transactions on Graphics (ToG)}, 37\penalty0 (6):\penalty0 1--15, 2018.

\bibitem[Hedman et~al.(2021)Hedman, Srinivasan, Mildenhall, Barron, and Debevec]{hedman2021baking}
Peter Hedman, Pratul~P Srinivasan, Ben Mildenhall, Jonathan~T Barron, and Paul Debevec.
\newblock Baking neural radiance fields for real-time view synthesis.
\newblock In \emph{Proceedings of the IEEE/CVF International Conference on Computer Vision}, pages 5875--5884, 2021.

\bibitem[Kalantari et~al.(2016)Kalantari, Wang, and Ramamoorthi]{kalantari2016learning}
Nima~Khademi Kalantari, Ting-Chun Wang, and Ravi Ramamoorthi.
\newblock Learning-based view synthesis for light field cameras.
\newblock \emph{ACM Transactions on Graphics (TOG)}, 35\penalty0 (6):\penalty0 1--10, 2016.

\bibitem[Laveau and Faugeras(1994)]{laveau19943}
Stephane Laveau and Olivier~D Faugeras.
\newblock 3-d scene representation as a collection of images.
\newblock In \emph{Proceedings of 12th International Conference on Pattern Recognition}, pages 689--691. IEEE, 1994.

\bibitem[Levin and Durand(2010)]{levin2010linear}
Anat Levin and Fredo Durand.
\newblock Linear view synthesis using a dimensionality gap light field prior.
\newblock In \emph{2010 IEEE Computer Society Conference on Computer Vision and Pattern Recognition}, pages 1831--1838. IEEE, 2010.

\bibitem[Levoy and Hanrahan(2023)]{levoy2023light}
Marc Levoy and Pat Hanrahan.
\newblock Light field rendering.
\newblock In \emph{Seminal Graphics Papers: Pushing the Boundaries, Volume 2}, pages 441--452. 2023.

\bibitem[Lindell et~al.(2021)Lindell, Martel, and Wetzstein]{lindell2021autoint}
David~B Lindell, Julien~NP Martel, and Gordon Wetzstein.
\newblock Autoint: Automatic integration for fast neural volume rendering.
\newblock In \emph{Proceedings of the IEEE/CVF Conference on Computer Vision and Pattern Recognition}, pages 14556--14565, 2021.

\bibitem[Max(1995)]{max1995optical}
Nelson Max.
\newblock Optical models for direct volume rendering.
\newblock \emph{IEEE Transactions on Visualization and Computer Graphics}, 1\penalty0 (2):\penalty0 99--108, 1995.

\bibitem[Mildenhall et~al.(2021)Mildenhall, Srinivasan, Tancik, Barron, Ramamoorthi, and Ng]{mildenhall2021nerf}
Ben Mildenhall, Pratul~P Srinivasan, Matthew Tancik, Jonathan~T Barron, Ravi Ramamoorthi, and Ren Ng.
\newblock Nerf: Representing scenes as neural radiance fields for view synthesis.
\newblock \emph{Communications of the ACM}, 65\penalty0 (1):\penalty0 99--106, 2021.

\bibitem[M{\"u}ller et~al.(2022)M{\"u}ller, Evans, Schied, and Keller]{muller2022instant}
Thomas M{\"u}ller, Alex Evans, Christoph Schied, and Alexander Keller.
\newblock Instant neural graphics primitives with a multiresolution hash encoding.
\newblock \emph{ACM Transactions on Graphics (ToG)}, 41\penalty0 (4):\penalty0 1--15, 2022.

\bibitem[Ng(2005)]{ng2005fourier}
Ren Ng.
\newblock Fourier slice photography.
\newblock In \emph{ACM Siggraph 2005 Papers}, pages 735--744. 2005.

\bibitem[Ng et~al.(2005)Ng, Levoy, Br{\'e}dif, Duval, Horowitz, and Hanrahan]{ng2005light}
Ren Ng, Marc Levoy, Mathieu Br{\'e}dif, Gene Duval, Mark Horowitz, and Pat Hanrahan.
\newblock \emph{Light field photography with a hand-held plenoptic camera}.
\newblock PhD thesis, Stanford university, 2005.

\bibitem[Oechsle et~al.(2021)Oechsle, Peng, and Geiger]{oechsle2021unisurf}
Michael Oechsle, Songyou Peng, and Andreas Geiger.
\newblock Unisurf: Unifying neural implicit surfaces and radiance fields for multi-view reconstruction.
\newblock In \emph{Proceedings of the IEEE/CVF International Conference on Computer Vision}, pages 5589--5599, 2021.

\bibitem[Penner and Zhang(2017)]{penner2017soft}
Eric Penner and Li Zhang.
\newblock Soft 3d reconstruction for view synthesis.
\newblock \emph{ACM Transactions on Graphics (TOG)}, 36\penalty0 (6):\penalty0 1--11, 2017.

\bibitem[Reiser et~al.(2023)Reiser, Szeliski, Verbin, Srinivasan, Mildenhall, Geiger, Barron, and Hedman]{reiser2023merf}
Christian Reiser, Rick Szeliski, Dor Verbin, Pratul Srinivasan, Ben Mildenhall, Andreas Geiger, Jon Barron, and Peter Hedman.
\newblock Merf: Memory-efficient radiance fields for real-time view synthesis in unbounded scenes.
\newblock \emph{ACM Transactions on Graphics (TOG)}, 42\penalty0 (4):\penalty0 1--12, 2023.

\bibitem[Seitz and Dyer(1995)]{seitz1995physically}
Steven~M Seitz and Charles~R Dyer.
\newblock Physically-valid view synthesis by image interpolation.
\newblock In \emph{Proceedings IEEE Workshop on Representation of Visual Scenes (In Conjunction with ICCV'95)}, pages 18--25. IEEE, 1995.

\bibitem[Seitz and Dyer(1996)]{seitz1996view}
Steven~M Seitz and Charles~R Dyer.
\newblock View morphing.
\newblock In \emph{Proceedings of the 23rd annual conference on Computer graphics and interactive techniques}, pages 21--30, 1996.

\bibitem[Shade et~al.(1998)Shade, Gortler, He, and Szeliski]{shade1998layered}
Jonathan Shade, Steven Gortler, Li-wei He, and Richard Szeliski.
\newblock Layered depth images.
\newblock In \emph{Proceedings of the 25th annual conference on Computer graphics and interactive techniques}, pages 231--242, 1998.

\bibitem[Shi et~al.(2014)Shi, Hassanieh, Davis, Katabi, and Durand]{shi2014light}
Lixin Shi, Haitham Hassanieh, Abe Davis, Dina Katabi, and Fredo Durand.
\newblock Light field reconstruction using sparsity in the continuous fourier domain.
\newblock \emph{ACM Transactions on Graphics (TOG)}, 34\penalty0 (1):\penalty0 1--13, 2014.

\bibitem[Shum and Kang(2000)]{shum2000review}
Harry Shum and Sing~Bing Kang.
\newblock Review of image-based rendering techniques.
\newblock In \emph{Visual Communications and Image Processing 2000}, pages 2--13. SPIE, 2000.

\bibitem[Sitzmann et~al.(2021)Sitzmann, Rezchikov, Freeman, Tenenbaum, and Durand]{sitzmann2021light}
Vincent Sitzmann, Semon Rezchikov, Bill Freeman, Josh Tenenbaum, and Fredo Durand.
\newblock Light field networks: Neural scene representations with single-evaluation rendering.
\newblock \emph{Advances in Neural Information Processing Systems}, 34:\penalty0 19313--19325, 2021.

\bibitem[Srinivasan et~al.(2017)Srinivasan, Wang, Sreelal, Ramamoorthi, and Ng]{srinivasan2017learning}
Pratul~P Srinivasan, Tongzhou Wang, Ashwin Sreelal, Ravi Ramamoorthi, and Ren Ng.
\newblock Learning to synthesize a 4d rgbd light field from a single image.
\newblock In \emph{Proceedings of the IEEE International Conference on Computer Vision}, pages 2243--2251, 2017.

\bibitem[Suhail et~al.(2022)Suhail, Esteves, Sigal, and Makadia]{suhail2022light}
Mohammed Suhail, Carlos Esteves, Leonid Sigal, and Ameesh Makadia.
\newblock Light field neural rendering.
\newblock In \emph{Proceedings of the IEEE/CVF Conference on Computer Vision and Pattern Recognition}, pages 8269--8279, 2022.

\bibitem[Thies et~al.(2018)Thies, Zollh{\"o}fer, Theobalt, Stamminger, and Nie{\ss}ner]{thies2018ignor}
Justus Thies, Michael Zollh{\"o}fer, Christian Theobalt, Marc Stamminger, and Matthias Nie{\ss}ner.
\newblock Ignor: Image-guided neural object rendering.
\newblock \emph{arXiv preprint arXiv:1811.10720}, 2018.

\bibitem[Vaish et~al.(2004)Vaish, Wilburn, Joshi, and Levoy]{vaish2004using}
Vaibhav Vaish, Bennett Wilburn, Neel Joshi, and Marc Levoy.
\newblock Using plane+ parallax for calibrating dense camera arrays.
\newblock In \emph{Proceedings of the 2004 IEEE Computer Society Conference on Computer Vision and Pattern Recognition, 2004. CVPR 2004.}, pages I--I. IEEE, 2004.

\bibitem[Verbin et~al.(2022)Verbin, Hedman, Mildenhall, Zickler, Barron, and Srinivasan]{verbin2022ref}
Dor Verbin, Peter Hedman, Ben Mildenhall, Todd Zickler, Jonathan~T Barron, and Pratul~P Srinivasan.
\newblock Ref-nerf: Structured view-dependent appearance for neural radiance fields.
\newblock In \emph{2022 IEEE/CVF Conference on Computer Vision and Pattern Recognition (CVPR)}, pages 5481--5490. IEEE, 2022.

\bibitem[Wang et~al.(2021)Wang, Wang, Genova, Srinivasan, Zhou, Barron, Martin-Brualla, Snavely, and Funkhouser]{wang2021ibrnet}
Qianqian Wang, Zhicheng Wang, Kyle Genova, Pratul~P Srinivasan, Howard Zhou, Jonathan~T Barron, Ricardo Martin-Brualla, Noah Snavely, and Thomas Funkhouser.
\newblock Ibrnet: Learning multi-view image-based rendering.
\newblock In \emph{Proceedings of the IEEE/CVF Conference on Computer Vision and Pattern Recognition}, pages 4690--4699, 2021.

\bibitem[Wilburn et~al.(2005)Wilburn, Joshi, Vaish, Talvala, Antunez, Barth, Adams, Horowitz, and Levoy]{wilburn2005high}
Bennett Wilburn, Neel Joshi, Vaibhav Vaish, Eino-Ville Talvala, Emilio Antunez, Adam Barth, Andrew Adams, Mark Horowitz, and Marc Levoy.
\newblock High performance imaging using large camera arrays.
\newblock In \emph{ACM SIGGRAPH 2005 Papers}, pages 765--776. 2005.

\bibitem[Wilburn et~al.(2001)Wilburn, Smulski, Lee, and Horowitz]{wilburn2001light}
Bennett~S Wilburn, Michal Smulski, Hsiao-Heng~Kelin Lee, and Mark~A Horowitz.
\newblock Light field video camera.
\newblock In \emph{Media Processors 2002}, pages 29--36. SPIE, 2001.

\bibitem[Wizadwongsa et~al.(2021)Wizadwongsa, Phongthawee, Yenphraphai, and Suwajanakorn]{wizadwongsa2021nex}
Suttisak Wizadwongsa, Pakkapon Phongthawee, Jiraphon Yenphraphai, and Supasorn Suwajanakorn.
\newblock Nex: Real-time view synthesis with neural basis expansion.
\newblock In \emph{Proceedings of the IEEE/CVF Conference on Computer Vision and Pattern Recognition}, pages 8534--8543, 2021.

\bibitem[Zhang et~al.(2021)Zhang, Luan, Wang, Bala, and Snavely]{zhang2021physg}
Kai Zhang, Fujun Luan, Qianqian Wang, Kavita Bala, and Noah Snavely.
\newblock Physg: Inverse rendering with spherical gaussians for physics-based material editing and relighting.
\newblock In \emph{Proceedings of the IEEE/CVF Conference on Computer Vision and Pattern Recognition}, pages 5453--5462, 2021.

\end{thebibliography}
}

% WARNING: do not forget to delete the supplementary pages from your submission
\clearpage
\setcounter{page}{1}
\maketitlesupplementary

\section{Proof for the Numerical Estimator (Sec. \ref{sec:method:taylor})}
\label{sec:suppl:proof}

Here we provide a step-by-step proof of the error-bound inequality for the numerical estimators.
Computing Taylor expansions, the second-order term of $\hat{C}(\rvr)$ is
\begin{equation}
    T^{(2)}\hat{C}(\rvr)
    = \frac{1}{2} \sum_{i=1}^N \lambda_i \Delta\rvh_i^t \nabla_\rvh^2 f_\rvc(\rvh_*,\rvd) \Delta\rvh_i
\end{equation}
The second-order term of $\hat{C}'(\rvr)$ is
\begin{align}
    &~ T^{(2)}\hat{C}'(\rvr) \\
    &= \frac{1}{2} \left(\sum_{i=1}^N\Delta\rvh_i\right)^t \nabla_\rvh^2 f_\rvc(\rvh_*,\rvd) \left(\sum_{i=1}^N\Delta\rvh_i\right) \\
    &= \frac{1}{2} \sum_{i=1}^N \sum_{j=1}^N \lambda_i\lambda_j \Delta\rvh_i^t \nabla_\rvh^2 f_\rvc(\rvh_*,\rvd) \Delta\rvh_j
\end{align}

Observe that for any symmetric matrix $\rmA$,
\begin{equation}
    |\rvh^t_1\rmA\rvh_2|
    = \langle \rvh_1, \rmA\rvh_2 \rangle
    \leqslant \|\rvh_1\| \|\rmA\rvh_2\|
    \leqslant \|\rvh_1\| \|\rmA\| \|\rvh_2\|
\end{equation}
The equality holds when $\rvh_1=\rvh_2$ and they align with the largest singular value of $\rmA$.
Specifically, when $\rvh_1=\rvh_2=\rvh$, we have $|\rvh^t\rmA\rvh| \leqslant \|\rmA\| \|\rvh\|^2$.

Let $\|\nabla_\rvh^2 f_\rvc(\rvh_*,\rvd)\|$ be the matrix Euclidean norm of the Hessian and denote $u_i = \|\nabla_\rvh^2 f_\rvc(\rvh_*,\rvd)\|^{1/2} \|\Delta\rvh_i\|$.
Therefore, with the triangle inequality, we have
\begin{align}
    \left| T^{(2)}\hat{C}(\rvr) \right|
    &\leqslant \frac{1}{2} \sum_{i=1}^N \lambda_i \left| \Delta\rvh_i^t \nabla_\rvh^2 f_\rvc(\rvh_*,\rvd) \Delta\rvh_i \right| \\
    &\leqslant \frac{1}{2} \sum_{i=1}^N \lambda_i \|\nabla_\rvh^2 f_\rvc(\rvh_*,\rvd)\| \|\Delta\rvh_i\|^2 \\
    &= \frac{1}{2} \sum_{i=1}^N \lambda_i u_i^2
\end{align}
Similarly, for $T^{(2)}\hat{C}'(\rvr)$ we have
\begin{align}
    &~ \left| T^{(2)}\hat{C}'(\rvr) \right| \\
    &\leqslant \frac{1}{2} \sum_{i=1}^N \sum_{j=1}^N \lambda_i\lambda_j \left| \Delta\rvh_i^t \nabla_\rvh^2 f_\rvc(\rvh_*,\rvd) \Delta\rvh_j \right| \\
    &\leqslant \frac{1}{2} \sum_{i=1}^N \sum_{j=1}^N \lambda_i\lambda_j \|\Delta\rvh_i\| \|\nabla_\rvh^2 f_\rvc(\rvh_*,\rvd)\| \|\Delta\rvh_j\| \\
    &= \left( \sum_{i=1}^N \lambda_i u_i \right)^2
\end{align}
This gives the (tight) upper bounds on the numerical errors
\begin{align}
    \left| T^{(2)}\hat{C} \right| &\leqslant
    U\left(\left| T^{(2)}\hat{C} \right|\right) =
    \frac{1}{2} \sum_{i=1}^N \lambda_i u_i^2 \\
    \left| T^{(2)}\hat{C}' \right| &\leqslant
    U\left( \left| T^{(2)}\hat{C}' \right|\right) =
    \frac{1}{2} \left( \sum_{i=1}^N \lambda_i u_i \right)^2
\end{align}
Finally, by Jensen's inequality, we have
\begin{equation}
     U\left(\left| T^{(2)}\hat{C} \right|\right) \geqslant
     U\left(\left| T^{(2)}\hat{C}' \right| \right)
\end{equation}
  
\section{Per-Scene Results (Sec. \ref{sec:exp})}
\label{sec:suppl:results}

The tables below show the per-category results for the experiments in Section \ref{sec:exp}).

\begin{table}[ht]
    \centering
    \setlength{\tabcolsep}{3pt}
    \begin{tabular}{l|cccccc}
    \toprule
         & Ball & Car & Coffee & Helmet & Teapot & Toaster \\
    \midrule
        \rowcolor[gray]{0.9}\multicolumn{7}{c}{Posi enc: \textbf{Sinusoidal}, view enc: \textbf{Sinusoidal}} \\
    \midrule
         Classic & 27.85 & 27.08 & \textbf{31.60} & 28.15 & 45.90 & \textbf{22.57} \\
         Ours & \textbf{28.61} & \textbf{27.31} & \textbf{31.60} & \textbf{28.53} & \textbf{45.74} & 22.31 \\
    \midrule
        \rowcolor[gray]{0.9}\multicolumn{7}{c}{Posi enc: \textbf{Sinusoidal}, view enc: \textbf{SH}} \\
    \midrule
         Classic & 27.92 & 26.97 & 31.09 & 27.99 & 45.19 & \textbf{21.81} \\
         Ours & \textbf{28.56} & \textbf{27.27} & \textbf{31.68} & \textbf{28.36} & \textbf{45.71} & 22.28 \\
    \midrule
        \rowcolor[gray]{0.9}\multicolumn{7}{c}{Posi enc: \textbf{TiledGrid}, view enc: \textbf{Sinusoidal}} \\
    \midrule
         Classic & 24.97 & 26.24 & \textbf{31.28} & 26.95 & 44.88 & 21.74 \\
         Ours & \textbf{26.76} & \textbf{26.46} & 31.21 & \textbf{27.61} & \textbf{45.14} & \textbf{21.94} \\
    \midrule
        \rowcolor[gray]{0.9}\multicolumn{7}{c}{Posi enc: \textbf{TiledGrid}, view enc: \textbf{SH}} \\
    \midrule
         Classic & 25.65 & \textbf{26.23} & 30.98 & 26.88 & 45.01 & 22.11 \\
         Ours & \textbf{27.05} & 26.22 & \textbf{31.12} & \textbf{27.63} & \textbf{45.19} & \textbf{22.27} \\
    \midrule
        \rowcolor[gray]{0.9}\multicolumn{7}{c}{Posi enc: \textbf{HashGrid}, view enc: \textbf{Sinusoidal}} \\
    \midrule
         Classic & 24.15 & 26.14 & \textbf{31.61} & \textbf{26.73} & 45.50 & 22.02 \\
         Ours & \textbf{25.58} & \textbf{26.62} & 31.54 & 26.70 & \textbf{45.86} & \textbf{22.23} \\
    \midrule
        \rowcolor[gray]{0.9}\multicolumn{7}{c}{Posi enc: \textbf{HashGrid}, view enc: \textbf{SH}} \\
    \midrule
         Classic & 24.15 & 27.00 & 31.31 & 27.36 & 45.70 & 22.23 \\
         Ours & \textbf{26.99} & \textbf{27.16} & \textbf{31.69} & \textbf{27.85} & \textbf{45.86} & \textbf{22.59} \\
    \bottomrule
    \end{tabular}
    \caption{Per-scene PSNR for different NeRF variations on the Shiny Blender dataset \cite{verbin2022ref}  (Tab. \ref{tab:exp_all_nerfs}, Sec. \ref{sec:exp:all_nerfs}).
    }
    \label{tab:suppl:exp_all_nerfs}
\end{table}

\begin{table}[ht]
    \centering
    \setlength{\tabcolsep}{3pt}
    \begin{tabular}{l|cc|cc|cc}
    \toprule
        & \multicolumn{2}{c|}{PSNR$\uparrow$} & \multicolumn{2}{c|}{SSIM$\uparrow$} & \multicolumn{2}{c}{LPIPS $\downarrow$} \\
        Method & Classic & Ours & Classic & Ours & Classic & Ours \\
    \midrule
        Ball & 24.15 & \textbf{26.99} & 0.793 & \textbf{0.796} & 0.256 & \textbf{0.239} \\
        Car & 27.00 & \textbf{27.16} & 0.915 & \textbf{0.916} & \textbf{0.083} & \textbf{0.083} \\
        Coffee & 31.31 & \textbf{31.69} & 0.962 & \textbf{0.963} & \textbf{0.130} & \textbf{0.130} \\
        Helmet & 27.36 & \textbf{27.85} & 0.916 & 0\textbf{.928} & 0.173 & \textbf{0.154} \\
        Teapot & 45.70 & \textbf{45.86} & 0.995 & \textbf{0.996} & 0.015 & \textbf{0.013} \\
        Toaster & 22.23 & \textbf{22.59} & 0.840 & \textbf{0.845} & 0.231 & \textbf{0.225} \\
    \bottomrule
    \end{tabular}
    \caption{Per-scene results on the Shiny Blender synthetic dataset \cite{verbin2022ref} (Tab. \ref{tab:exp_refnerf}, Sec. \ref{sec:exp:more_comparisons}).
    }
    \label{tab:suppl:exp_refnerf}
\end{table}

\begin{table}[ht]
    \centering
    \setlength{\tabcolsep}{3pt}
    \begin{tabular}{l|cc|cc|cc}
    \toprule
        & \multicolumn{2}{c|}{PSNR$\uparrow$} & \multicolumn{2}{c|}{SSIM$\uparrow$} & \multicolumn{2}{c}{LPIPS $\downarrow$} \\
        Method & Classic & Ours & Classic & Ours & Classic & Ours \\
    \midrule
        Chair & 35.70 & \textbf{35.79} & \textbf{0.986} & \textbf{0.986} & 0.022 & \textbf{0.021} \\
        Drums & 25.11 & \textbf{25.60} & 0.931 & \textbf{0.933} & 0.091 & \textbf{0.080} \\
        Ficus & \textbf{33.87} & 33.82 & 0.981 & \textbf{0.982} & 0.025 & \textbf{0.024} \\
        Hotdog & \textbf{37.45} & 37.39 & \textbf{0.981} & \textbf{0.981} & \textbf{0.036} & \textbf{0.036} \\
        Lego & 35.70 & \textbf{35.79} & \textbf{0.980} & \textbf{0.980} & \textbf{0.025} & \textbf{0.025} \\
        Materials & 29.60 & \textbf{29.73} & 0.948 & \textbf{0.951} & 0.069 & \textbf{0.066} \\
        Mic & \textbf{36.73} & 36.58 & \textbf{0.991} & \textbf{0.991} & \textbf{0.014} & \textbf{0.014} \\
        Ship & 30.53 & \textbf{30.63} & \textbf{0.888} & 0.887 & 0.148 & \textbf{0.144} \\
    \bottomrule
    \end{tabular}
    \caption{Per-scene results on the Blender synthetic dataset \cite{mildenhall2021nerf}  (Tab. \ref{tab:exp_blender}, Sec. \ref{sec:exp:more_comparisons}).
    }
    \label{tab:suppl:exp_blender}
\end{table}

\begin{table}[ht]
    \centering
    \setlength{\tabcolsep}{3pt}
    \begin{tabular}{l|cc|cc|cc}
    \toprule
        & \multicolumn{2}{c|}{PSNR$\uparrow$} & \multicolumn{2}{c|}{SSIM$\uparrow$} & \multicolumn{2}{c}{LPIPS $\downarrow$} \\
        Method & Classic & Ours & Classic & Ours & Classic & Ours \\
    \midrule
        CD & 31.41 & \textbf{31.50} & 0.935 & \textbf{0.937} & 0.091 & \textbf{0.090} \\
        Crest & 21.68 & \textbf{21.69} & \textbf{0.697} & 0.696 & \textbf{0.172} & 0.174 \\
        Food & \textbf{22.96} & 22.93 & \textbf{0.739} & \textbf{0.737} & \textbf{0.255} & 0.257 \\
        Giants & 26.37 & \textbf{26.38} & \textbf{0.840} & \textbf{0.840} & \textbf{0.173} & 0.175 \\
        Lab & 30.28 & \textbf{30.43} & 0.924 & \textbf{0.926} & 0.108 & \textbf{0.105} \\
        Pasta & \textbf{21.85} & 21.77 & 0.757 & \textbf{0.758} & 0.250 & \textbf{0.247} \\
        Seasoning & \textbf{28.30} & 28.28 & \textbf{0.850} & 0.849 & \textbf{0.229} & \textbf{0.229} \\
        Tools & \textbf{27.88} & 27.87 & \textbf{0.919} & \textbf{0.919} & 0.157 & \textbf{0.155} \\
    \bottomrule
    \end{tabular}
    \caption{Per-scene results on the Shiny real scene dataset \cite{wizadwongsa2021nex} (Tab. \ref{tab:exp_shiny}, Sec. \ref{sec:exp:more_comparisons}).
    }
    \label{tab:suppl:exp_shiny}
\end{table}

\section{Additional Explanations (Sec. \ref{sec:exp:more_comparisons})}
\label{sec:suppl:explana}

\begin{figure*}[t]
    \centering
    \includegraphics[width=.8\linewidth]{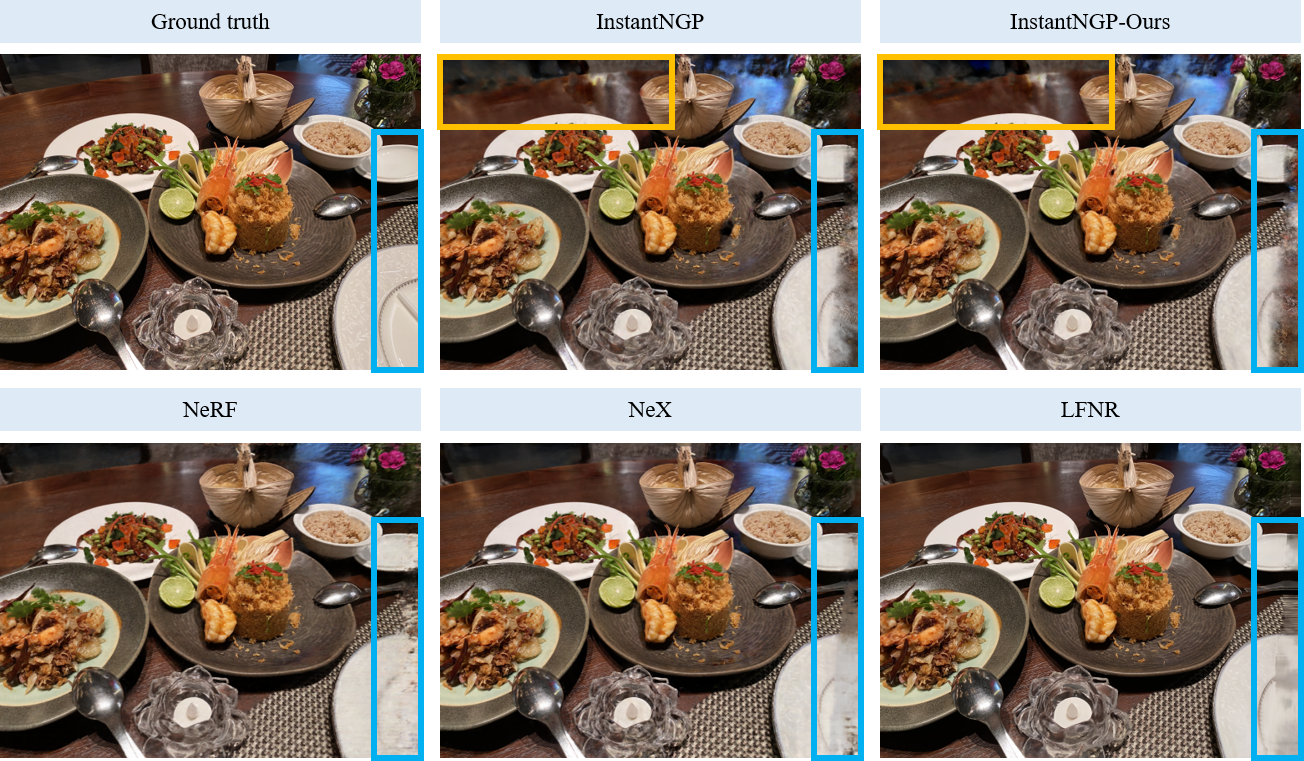}
    \caption{"Food" scene from the Shiny dataset \cite{wizadwongsa2021nex}.
    Both our method and classic NeRF struggle in the regions far from the camera ({\color{my_yellow}yellow boxes}), which is irrelevant to the rendering quality view-dependent effects.
    Moreover, in certain regions ({\color{my_blue}blue boxes}), information from the other views is insufficient for inferring the radiance at this view, and all volume rendering or light field rendering methods fail.
    }
    \label{fig:suppl:shiny_explanation}
\end{figure*}

As stated in Section \ref{sec:exp:more_comparisons}, compared to the two object-centric datasets (Blender \cite{mildenhall2021nerf} and Shiny Blender \cite{verbin2022ref}), our our improvements are less significant on the Shiny scene dataset \cite{wizadwongsa2021nex}.
Based on our observations, this is partly because of challenges in complex scene representations dominating the evaluation metrics.
Figure \ref{fig:suppl:shiny_explanation} shows an example of it with explanations.

\end{document}